# A Review on Explainable Artificial Intelligence for Healthcare: Why, How, and When?

Subrato Bharati, M. Rubaiyat Hossain Mondal, *Member, IEEE*, and Prajoy Podder

*Abstract*—Artificial intelligence (AI) models are increasingly finding applications in the field of medicine. Concerns have been raised about the explainability of the decisions that are made by these AI models. In this article, we give a systematic analysis of explainable artificial intelligence (XAI), with a primary focus on models that are currently being used in the field of healthcare. The literature search is conducted following the preferred reporting items for systematic reviews and meta-analyses (PRISMA) standards for relevant work published from 1 January 2012 to 02 February 2022. The review analyzes the prevailing trends in XAI and lays out the major directions in which research is headed. We investigate the why, how, and when of the uses of these XAI models and their implications. We present a comprehensive examination of XAI methodologies as well as an explanation of how a trustworthy AI can be derived from describing AI models for healthcare fields. The discussion of this work will contribute to the formalization of the XAI field.

*Impact Statement* — XAI is emerging in various sectors and the literature has recognized numerous concerns and problems. Security, performance, vocabulary, evaluation of explanation, and generalization of XAI in the field of healthcare are the most significant problems recognized and discussed in the literature. The paper has identified a number of obstacles associated with XAI in the healthcare area, including System Evaluation, Organizational, Legal, socio-relational, and Communication issues. Although a few studies in the literature have proposed some answers for the stated problems, others have characterized it as a future research gap that must be filled. In healthcare, the organizational difficulty posed by XAI can be mitigated via AI deployment management plans. The doctor's understanding of how patients would interpret the system and double-checking health facts with patients might help alleviate communication problems. The socio-organizational issue or difficulty can be overcome through patient instruction that facilitates the use of artificial intelligence. The applications of AI in healthcare, including diagnosis and prognosis, medication discovery, population health, healthcare organization, and patient-facing apps, have been reviewed critically. This article addresses a number of AI's limitations, including the difficulties of authenticating AI's output. Active collaboration between physicians, theoretical researchers, medical imaging specialists, and medical image analysis specialists will be crucial for the future development of machine learning and deep learning techniques. This paper contributes to the body of knowledge in identifying different techniques for XAI, in determining the issues and difficulties associated with XAI, and finally in reviewing XAI in healthcare.

## I. INTRODUCTION

Advances in Artificial Intelligence (AI) have taken place in recent years, with the technology becoming more complicated, more autonomous, and more sophisticated. As data grows exponentially, so does the speed at which computer power advances. This combination gives a massive boost to the development of AI. People's lives have already been impacted by AI's practical accomplishments in a range of fields (e.g., voice recognition, self-driving cars, and recommendation systems) [1-4]. AI can improve people's well-being and health by increasing clinicians' diagnostic work, highlighting prevention options, and delivering tailored treatment suggestions on electronic health records (EHRs). Despite the employment of several useful technologies in health care [5, 6], AI is not widely deployed [7, 8]. As a result, AI, machine learning (ML) and deep learning (DL) algorithms continue to be mysterious in many cases [9-11]. Decision-making processes are generally hard to explain. Moral and practical challenges arise. Without understanding AI diagnoses, it is impossible to tell if diagnostic disparities between patients reflect diagnostically important variations or bias, diagnostic errors, or under/over-diagnosis.

A number of governmental and commercial organizations have recently released recommendations and guidelines for the appropriate use of AI in response to these issues. The study, however, reveals a considerable discrepancy in the alleged transparency characteristics. Many critical difficulties in medicine may be resolved with the use of AI. Numerous cases have gained pace in recent years as a result of significant research on automated prognosis, diagnosis, testing, and medicine formulation [12-14]. Along with genetic data, biosensors, and medical imaging, these sources create a large quantity of data [15-17]. A critical objective of AI in medicine is to personalize medical decisions, health practices, and medications for unique people. Nevertheless, the current state of AI in medicine has been described as "strong on promise but rather lacking in evidence and demonstration". Real-world clinical settings have been tested to detect wrist fractures,

Subrato Bharati was with the Institute of Information and Communication Technology, Bangladesh University of Engineering and Technology, Dhaka-1205, Bangladesh, and now with the Department of Electrical and Computer Engineering, Concordia University, Montreal, QC, Canada. (e-mail: subratobharati1@gmail.com) (Corresponding Author).

M. Rubaiyat Hossain Mondal is with the Institute of Information and Communication Technology, Bangladesh University of Engineering and Technology, Dhaka-1205, Bangladesh (e-mail: rubaiyat97@iict.buet.ac.bd).

Prajoy Podder is with the Institute of Information and Communication Technology, Bangladesh University of Engineering and Technology, Dhaka-1205, Bangladesh (e-mail: prajoypodder@gmail.com).

diabetic retinopathy, cancer metastases, histologic breast, congenital cataracts, and colonic polyps; however, many of the systems that have been demonstrated to be equivalent to or superior to experts in experimental settings have demonstrated high false-positive rates in real-world clinical settings [18]. Additionally, prejudice, security, privacy, and a deficit of transparency exist, as do trust, fairness, informativeness, transferability, and causality [19]. Numerous critical difficulties in medicine may be resolved with the use of AI [3, 20-25]. Together with medical imaging, biosensors, genetic data, and electronic medical records, these sources create a large quantity of data [18]. To make a precise diagnosis in precision medicine, doctors need substantially more information than a simple binary prediction [19, 27]. ML in medicine has grown more concerned with explainability, interpretability, and openness [19]. There is sufficient evidence that ML-based methods are helpful; however, their widespread adoption is unlikely until these issues are solved, most likely by providing systems with adequate explanations for their judgments [28]. Because of this, it is difficult to develop generalizable solutions that can be used in a variety of contexts. For example, various applications often have distinct interpretability and explainability demands [19, 26-28].

Explainable AI (XAI) and interpretable AI (IAI) are not the same concepts. Although some studies used the terms XAI and IAI interchangeably, the difference and relations between these two are described in the literature in multiple ways. This is presented in the following. XAI outlines why the choice was made but not how the decision was reached. The term IAI outlines how the choice was made but not why the criteria used were reasonable [171]. Explainability means that a ML model and its results can be explained in a way that a person can understand. XAI permits the explanation of learning models and focuses on why the system made a given decision, analyzing its logical paradigms. On the other hand, the interpretability of ML enables users to comprehend the results of the learning models by revealing the rationale for its decisions [171, 172]. According to some literature [173], explainability is the concept that a ML model and its findings may be communicated to a person in words that make sense. In one work, Adadi and Berrada et al. [173] described that interpretable systems are explainable if their processes can be comprehended by humans, demonstrating that the idea of explainability is closely related to that of interpretability. Nonetheless, Gilpin et al. The authors of [174] noted that interpretability and fidelity are both essential components of explainability. They stated that a good explanation must be human-comprehensible (interpretability) and accurately characterize model behavior over the whole feature space (fidelity). Interpretability is necessary to handle the social interaction of explainability, whereas fidelity is necessary to assist in confirming other model prerequisites or discovering new explanations. In other words, the fidelity of an explanation relates to how reliably and exactly the behavior of a model is explained. So, an explanation is explainable if it is easily understood by humans and accurately explains the model.

Figure 1 depicts the literature search conducted for this article in accordance with the PRISMA (preferred reporting items for systematic reviews and meta-analyses) standards. PRISMA guidelines and the expanded version of PRISMA scoping reviews were used to create this scoping review (PRISMA-ScR). The number of records considered, records included, and records deleted for further evaluation are shown in a flow diagram in Figure 1. After conducting a thorough literature review, all relevant research publications were found. The dates 1 January 2012 to 02 February 2022 were used as the unit of measurement. Records in English were considered, while those in other languages were eliminated. Records were examined to see whether they met the requirements of the PICO (patient, intervention, comparator, and outcome) study. Initial attention was given to studies that met all four of the following criteria: Using XAI for screening or diagnosing patients, as well as comparing it to conventional approaches, was shown to be effective. Two Boolean operators, OR and AND, were used to find the most important terms. Many terms were searched for, including XAI, explainable AI, interpretable AI, medical medicine, medical, etc. In addition, the publications' titles, abstracts, and keywords were all taken into account while doing the search. No papers on XAI fundamental scientific, epidemiology, or clinical aspects were reviewed. It was then decided to eliminate any content that covered the same topic twice. Extracting information and synthesizing data elements such as research objectives, and performance outcomes were the next step in the process.

Elsevier (ScienceDirect), IEEE, MDPI, SpringerNature, Bentham Science, Wiley, Medrxiv, Arxiv and many more were among the reputable publishers and preprint servers that yielded a total of 17,300 publications on XAI for medical care or medicine purposes. The Appendix contains the search syntax used in this article. Out of a total of 17,300 documents, 16,850 were manually excluded because they did not correspond to our primary research area. Therefore, the healthcare or medicine or medical treatment application submitted by XAI did not match the necessary criteria. In total, 450 papers were evaluated; however, only 195 were chosen for inclusion in the research and the report. From the 195 papers assessed, 110 were chosen for inclusion in the summary of the findings as these reported the performance measurements and the performance outcomes of XAI for healthcare or medicine. To put it another way, any of the following problems have been covered by any of these 110 publications. Our classification of XAI in healthcare and medicine into five categories synthesized by the method of the authors of [135], which includes (1) dimension reduction using XAI, (2) feature selection using XAI, (3) attention mechanism using XAI, (4) knowledge distillation using XAI, and (5) surrogate representations using XAI, is presented in this review. Although some recent studies have reviewed XAI, ongoing surveys and studies are necessary because the field of XAI in healthcare is always changing. With this in consideration, this review paper extends the findings reported by previous survey papers in the field of XAI [19, 26, 73, 80, 81, 87, 195].

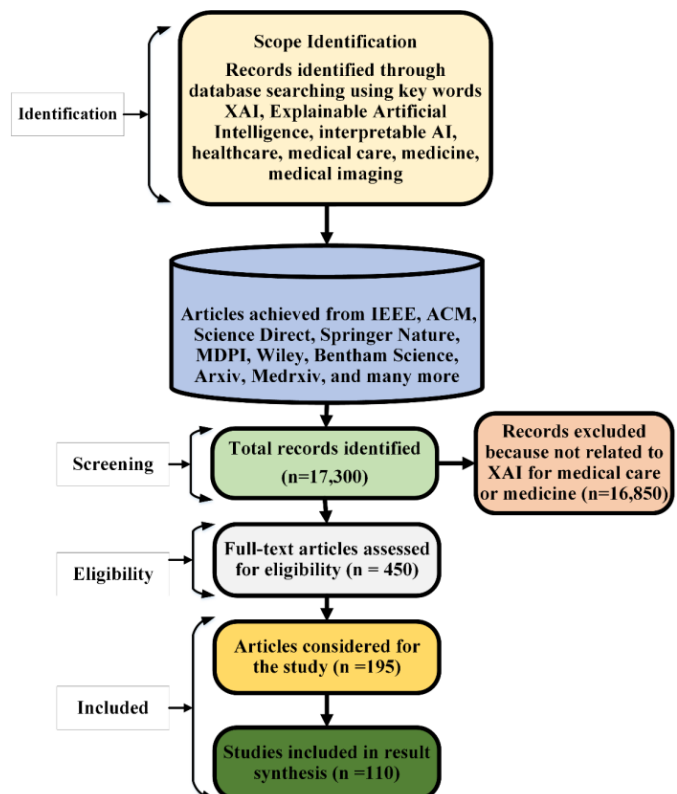

Fig. 1. PRISMA guideline for manuscript selection for this review

The research mainly focuses on the emerging field of XAI and its challenges in health care. This paper also reports some IAI models, as some literature uses the terms XAI and IAI interchangeably. The main contributions of this paper is in presenting the answers to the following questions:
- Why use XAI in health care and medicine, and what problems cannot be solved with standard AI but can only be solved using XAI?
- What are the best ways to accomplish our XAI?
- When is the best time to perform XAI?

The remaining sections are organized as follows. Background information is provided in Section II of this review work. In Section III, we go through the benefits of utilizing XAI as well as the appropriate times to do so. Section IV presents the XAI methods describing how it is used. In Section V, several different XAI application situations are broken down and examined. The overview of various XAI aspects and the most recent research developments in XAI are covered in the next part, Section VI. In Section VII, the final thoughts are presented.

## II. BACKGROUND

By mimicking human mental processes, AI presents a paradigm shift in healthcare data-driven by increasing data availability in healthcare as well as fast developments in mathematical methodologies. This article examines healthcare AI applications that are currently being developed. General AI techniques for healthcare are discussed in the literature [29], [30], [31], while medical image analysis is described in [32-34]. The consequences of AI have been disputed in the medical literature [35-37]. An AI-powered device [38] may assist in clinical decision-making by delivering recent medical information from conferences, manuals, and journals. AI systems may also assist in eliminating therapeutic and medical errors that are inescapable (such as those that are more repeatable and objective) in human clinical practice [38-40]. An AI system may also use data from a large number of patients to provide real-time insights for outcome estimation and health risk warning. For decades, rule-based algorithms have made great strides and have been used to interpret ECGs, detect disorders, choose the best treatment option for each patient, and support clinicians in establishing diagnostic theories and hypotheses in difficult instances of patients. Because they need the explicit specification of necessitating human-authored updates, decision rules and rule-based systems are both time and cost-intensive to construct. Thus, the effectiveness of the structures is limited by the extent to which past medical knowledge encompasses higher-order relationships between material published in different specialties [20, 41]. Likewise, it was challenging to include a strategy that combines probabilistic and deterministic reasoning methods to minimize the suitable recommended therapy, prioritize medical theories, and psychological environment [42-43]. In contrast to the 1st generation of AI programs, which depended only on medical information curation by specialists as well as the construction of state-of-the-art AI research, strict decision rules have utilized ML methodologies, which can be used to explain complex relationships [44]. By studying the patterns produced from all the labeled input-output pairings, ML algorithms learn to offer the appropriate output for a provided input in new cases [45]. Algorithms for supervised ML are designed to find the model parameters that minimize the discrepancy from their predicted impacts in their training cases to the observed effects in these situations, in the hopes that the contexts established will be transferable to other datasets. The test dataset may be used to determine the generalizability of the model. For an unlabeled collection of data, unsupervised learning may help to uncover sub-clusters of the main dataset or recognize outliers in the dataset. As a result, it is worth noting that low-dimensional depictions of labeled datasets may be recognized in a supervised way. It is possible to design applications of AI that allow for the earlier study of unrecognized patterns in data apart from the requirement to specify decision-making rules for every task or take into consideration sophisticated relations between input elements. Because of this, ML has become the most commonly used approach for constructing AI applications [2, 46-48].

The current resurgence of AI is largely due to the active application of DL, which comprises training a deep neural network on large datasets [49]. Fortunately, the processing resources necessary for DL have been enabled by recent improvements in computer processor design [50]. Due to the establishment of a range of large-scale research [51, 52], data collection platforms [53], these algorithms have only recently become available. A wide range of DL-based models has been created for prediction and diagnosis, image classification, biomarker identification, genomic interpretation, and monitoring via automated robotic surgery wearables and life-logging devices to improve digital healthcare [54]. Due to the quick expansion of AI, healthcare data may be used to construct DL-based strong algorithms that can automate diagnosis and provide a more precise technique of medicine and health by customizing pharmaceuticals and aiming for services in a dynamic and timely manner. A non-exhaustive list of potential uses is shown in Figure 2. Medical AI malpractice claims will continue to get precise instructions from the legal system as to which agency is liable if they arise. As part of the AI framework, healthcare decisions are made in part by healthcare practitioners with malpractice insurance [31, 55, 56]. It is important to minimize bias while maximizing patient satisfaction with the implementation of AI modules in these areas [57, 58]. It is shown in Figure 2 that an auxiliary advancement of XAI might possibly solve the problem of limited sample learning by removing clinically insignificant characteristics. DL models, on the other hand, may often provide results that are beyond the comprehension of non-experts. A DL model may have millions of parameters, which might make understanding what the machine monitors in clinical health data, such as CT scans, challenging [3, 48 59].

unsolved [66]. It can be noted that risk and obligations are only two examples of concerns that are not recognized in other domains when it comes to clinical interpretations [30, 67]. Letting AI algorithms make life-or-death medical choices without providing enough transparency and accountability is irresponsible [68]. A number of recent studies [6, 69-73] have examined the issue of explainability in medical AI. COVID-19 classification and detection, emotion analysis, chest radiography, as well as the relevance of interpretability in the medical industry have all been studied in detail [74-75]. Figure 3 depicts an overview of XAI system. Some research works rely mostly on keeping the interpretability of AI models intact while enhancing their performance via optimization and refining procedures [76-79]. The authors of [80] studied XAI in healthcare, whereas the authors of [81] studied XAI for electronic health records. Therefore, the purpose of this study is to provide a synthesis of the categories of XAI in healthcare and medicine by analyzing them in terms of dimension reduction, feature selection, attention mechanism, knowledge distillation, and surrogate representations utilizing XAI.

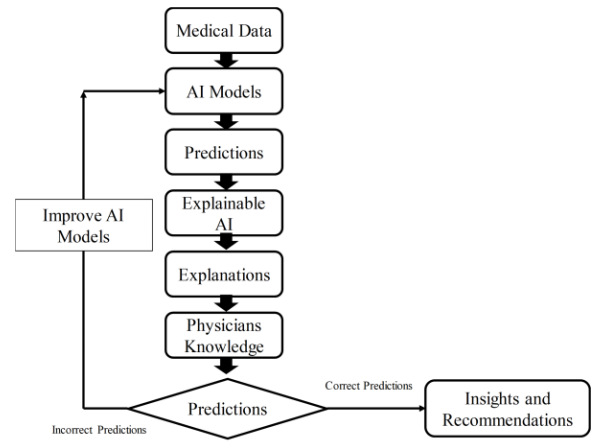

Fig. 3. An overview of XAI system step by step

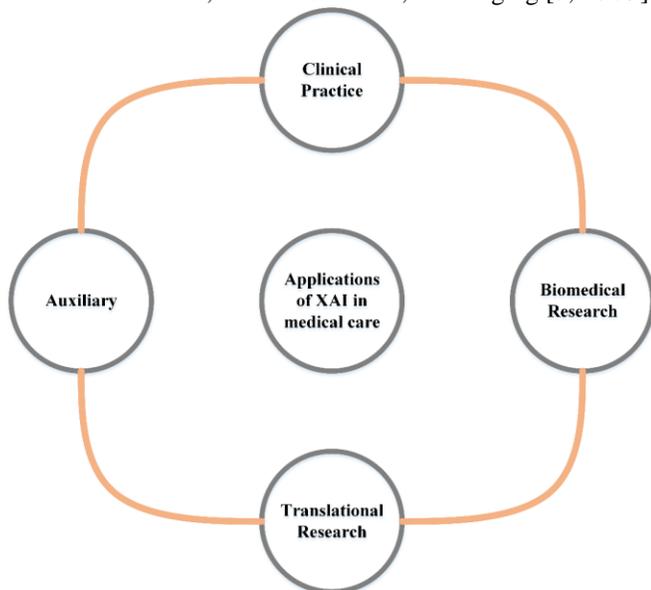

Fig. 2. Applications of XAI

Researchers have noted specifically that being a major restriction for AI in dermatology due to its inability to perform a tailored assessment by doctors [60]. There must be proof that a high-performance DL algorithm genuinely recognizes the correct region of the image and does not overemphasize insignificant discoveries. In recent years, new ways to describe AI models have been created, including methods for visualizing them. Attention maps [61], class activation maps [62-63], salience maps [64], and occlusion maps [65] are some of the most frequently used levers in the field of visual perception. Because the result is a picture, localization and segmentation techniques are easier to understand. A key problem in understanding deep neural network (DNN) based models trained on non-imaging medical data other than images is yet

### III. REASONS FOR USING XAI

Recently, AI methods like DL have played a revolutionary role in healthcare, particularly in the areas of diagnostics and surgery. Some deep-learning-based diagnosis jobs are even more accurate than those performed by doctors. However, the DL model's lack of transparency makes it difficult to explain their results and implement them in clinical settings. A growing number of experts in the intersection of AI and healthcare have concluded that an AI model's capacity to be explained to humans is more important than the model's accuracy when it comes to its practical application in a clinical setting. It is important to understand how medical AI applications work before they are adopted and used. As a result, there is an incentive to conduct a survey of the medical XAI, and XAI is necessary for the acceptability of AI applications in medicine.

### IV. XAI METHODS

This section describes the methods of XAI described in the literature [12, 84, 85-87]. For example-based, attribution-based, and model-based explanations, the qualities of explainability, such as interpretability characterized by simplicity and clarity, as well as fidelity characterized by soundness and completeness, are evaluated in this section.

Figure 3 depicts an overview of XAI system. Some research works rely primarily on keeping the interpretability of AI models intact while enhancing their performance via optimization and refining procedures [76-79]. The work in[80] and [81] studied XAI for healthcare and electronic health records, respectively. The works of [82] and [83] focus on specialized applications, such as digital pathology.

*A. Analyzing explanations based on models*

Model-based descriptions meet the completeness criteria since they offer enough information to calculate the outcome for a given input. Fidelity can be quantified by looking at the proportion of predictions that are consistent with the task model's post-hoc explanations. Global explanations, which give a single reason for the whole model, are more likely to meet the criterion of clarity. Because of this, an example-based, attribution-based, or global model-based approach is the best option for providing clarity. Using rule overlap and feature space coverage, the researchers of [132] established a measure of unambiguity that may be applied to this particular scenario. On the other hand, local explanations are plagued by the issue of the great variation in explanations for the same or comparable cases. Local model-based explanations' clarity could not be measured using any existing metric. As a measure of model interpretability and as an estimate of model complexity [85, 94, 133], model size is often utilized. For example, the number of features (the number of rules, non-zero weights utilized in splits), as well as the complexity of relationships (the depth of the tree, the length of rules, interactions), are examples of metrics, however, they are model dependent. For example, the authors of [91] suggested assessing model complexity by counting the number of arithmetic and boolean operations required for the given input. They also examined how complexity affected model accuracy. Molnar et al. [95] calculated the level of model complexity by looking at the number of features, the intensity of interactions, and complexity of the principal effects.

*B. Analyzing explanations based on attribution*

Attribution-based descriptions only give a limited description of the model, and hence do not meet the requirement of completeness. There were no metrics available to measure the completeness of an explanation based on attribution. However, the majority of attribution-based explanation evaluation approaches may be utilized to quantify soundness. The effectiveness of the attribution technique may be empirically evaluated. In certain cases, this is accomplished via the use of ground truth created artificially [66, 96]. Other approaches use perturbation analysis to measure performance decline [97,98]. It is the goal of axiomatic evaluations to establish what an ideal attribution technique should accomplish, and then to assess whether this has been done. In order to see whether an explanation is affected by a nonsensical change in the input data, the authors of [98] offer a test. Sensitivity and implementation invariance are also examined by Sundararajan, Taly, and Yan [99]. When a feature's non-zero attribution leads to various forecasts for a given baseline, they designate an attribution approach as sensitive. Implementation invariant means that an attribution method always returns the same attribution for functionally similar models (i.e., the similar output is produced by the similar input). Explanation continuity and selectivity are essential qualities discussed by Montavon et al. [100]. The authors of [101] report that an attribution approach meets sensitivity-N when the total of the attributions for any subset of cardinality N features is equal to the change in the output produced by eliminating the features in the subset.

*C. Analyzing explanations based on example*

An evaluation methodology for describing examples may be insufficient because there are fewer tools for obtaining influential cases and prototypes. Global examples are clear and example-based explanations are efficient if the case is understandable. Other conditions were not met, and no assessment techniques were found. Table I summarizes the results of this section.

TABLE I
AI APPROACHES THAT CAN BE EXPLAINED IN TERMS OF THE CHARACTERISTICS OF EXPLAINABILITY

| Method | Explanation type | Scope | Fidelity | | Interpretability | |
|---|---|---|---|---|---|---|
| | | | **Soundness** | **Completeness** | **Parsimony** | **Clarity** |
| Post-hoc explanation | Attribution | Local | General quantitative metric | Unavailable matrices | Satisfied if an instance or feature is human-intelligible. | General quantitative metric |
| | Attribution | Global | General quantitative metric | Unavailable matrices | Satisfied if an instance or feature is human-intelligible. | Satisfied |
| | Model | Local | General quantitative metric | Satisfied | General quantitative metric | Unavailable matrices |
| | Model | Global | General quantitative metric | Satisfied | General quantitative metric | If the model is incapable of providing various rationales, the model is satisfied. |
| | Example | Local | Unavailable matrices | Unavailable matrices | Satisfied if an instance or feature is human-intelligible. | Unavailable matrices |
| | Example | Global | Unavailable matrices | Unavailable matrices | Satisfied if an instance or feature is human-intelligible. | Satisfied |

## V. APPLICATIONS SCENARIOS

This section presents the current and potential application scenarios of XAI.

### A. Dimension reduction using XAI

Deciphering AI models by depicting their most essential aspects is regularly and customarily accomplished via the use of dimension reduction methods such as Laplacian Eigenmaps [102, 103], independent component analysis (ICA) [104], and principal component analysis (PCA) [105]. The ideal dimensions of the input characteristics may be used to estimate side effects pharmacologically by merging multi-label as well as k-nearest-neighbor approaches [106]. Unsupervised classification of 1H MR spectroscopy brain tumor data was improved using a nonlinear dimension reduction approach suggested by the authors of [107], which used Laplacian Eigenmaps to identify the most significant characteristics. Patients with stage-one lung cancer were categorized by the authors of [108] using a supervised learning approach, which defined the most informative cases. A hybrid technique for dimension reduction, integrating pattern recognition and regression analytics, was presented to recognize a collection of "exemplars" in order to build a "dense data matrix." In the end, they used the most accurate instances in their model. On the basis of their expertise in the field of cell-type-specific enhancers, Kim and his colleagues [109] used DL to identify which traits were most significant and ranked them in order of importance in the model. Researchers of [110] presented a sparse DL approach with pathway-associated to investigate gene pathways as well as their correlations in brain tumor patients. Using dimension reduction techniques, it is possible to understand the model's underlying behavior by reducing the information to a tiny subset.

### B. Rule extraction and knowledge distillation using XAI

Learning a student model that is easy to understand with a trainer model that is difficult to explain is called knowledge distillation. This is a special kind of model-specific XAI that involves extracting information from a complex model and simplifying it. Dimension reduction and tree regularization [111] or a combination technique of model compression and dimension reduction [112] may be used to achieve this goal [113]. Hinton et al. [114] are among the researchers who have studied this strategy for some time, however, recent advancements in XAI and IAI [115-117] have given it new life. In addition to knowledge distillation, rule extraction is a commonly applied XAI technology that is well-suited for usage in digital healthcare. For example, the Model Understanding via Subspace Explanations (MUSE) technique [118] has been created to characterize the projections of the global model by examining several subgroups of instances. For example, rule sets or decision sets have been explored for interpretability [89]. In the context of XAI and IAI, mimic-learning is a fundamental strategy for distilling information that uses gradient-boosting trees to learn interpretable structures and make the original model understandable. An interpretable model for ICU outcomes, such as death or ventilator use, was built using the distilled information from this technique. An explainable description of the risk of pneumonia-related mortality was proposed by Caruana et al. in the work of XAI and IAI [119]. Bayesian rule lists were introduced by Letham et al. as IAI [120] and claimed to be able to predict strokes. A model induction technique developed by the authors of [121] applied visualization to create rules by the probability of a complex algorithm in a variety of tasks, including the classification of diabetes and the detection of breast cancer. By converting EHR occurrences into embedded clinical principles, DL was used by the authors of [122] to break dynamic connections between potential risk factors and hospital readmission for patients. An XAI system based on rule extraction was created by Davoodi and Moradi [123] to predict ICU mortality, while the authors of [124] employed a comparable XAI technique for Alzheimer's disease diagnosis. Textual reasoning was applied to the LSTM-based breast mass categorization by the authors of [125]. Argumentation theory was used in [126], where the XAI algorithm was applied to the training process for risk and stroke prediction.

### C. Feature importance or selection using XAI

Using feature importance as a tool, researchers have been able to better describe the properties and importance of extracted features, as well as the relationships between those features and the outcomes they predict [12, 113, 127]. The authors of [128] employed weights in features to identify the top 10 extractable characteristics or features for predicting mortality in the critical care unit. The researchers of [129] developed an algorithm for clinically relevant and risk-calculating prostate cancer [130] by estimating the importance of features by applying Shapley's value. The relevance of a feature may be gauged by doing a sensitivity analysis on the extracted features; the more sensitive the result, the more significant the feature. To identify the most important elements of a microbiota-based diagnostic job, Eck et al. [131] reduced the features and tested their influence on model performance. The other authors in [132] used a backpropagation-based technique to achieve interpretability by using DL important features (DeepLIFT). As a result of the backpropagation method, it is possible to determine the importance of a feature. By calculating the nucleotide contribution score for each nucleoside, the authors of [133] created interpretable DL that are IAI models for predicting splice sites using the DeepLIFT approach. Layer-wise relevance propagation (LRP) [134], guided backpropagation (GBP) [135], SHapley Additive exPlanations (SHAP) [136], DeepLIFT [132], and other XAI and IAI methods, were compared for diagnosing ophthalmic [137]. By extracting feature significance, XAI not only provides an explanation of features but also reflects their relative relevance to clinical interpretation. However, numerical weights can be either misinterpreted or difficult to grasp.

### D. Attention mechanism in XAI

Global and local attention mechanisms use all words to produce context, and self-attention mechanisms use several mechanisms simultaneously to locate every link between words [138]. There is evidence that the attention mechanism aids in both the improvement of interpretability and the advancement of visualization technology [139]. It has been shown in the patients of ICU utilizing the attention mechanism by the work of [140] that some input characteristics have a greater impact on clinical event predictions than others. Patients in the intensive care unit (ICU) may be assessed for severity using an interpretable acuity score framework developed by Shickel et al. [141] that makes use of attention and DL-based evaluation on sequential organ failure. HIV genome integration sites were accurately predicted by Hu and colleagues [142] using mechanistic theories. In addition, using a similar XAI strategy, the authors of [143] constructed a system for learning how to represent EHR data in a way that might explain the relationship between clinical results for individual patients. The Reverse Time Attention Model (RETAIN) was built by Choi et al. [144] and comprised two sets of attention weights: one at the visit level to account for the effect of each visit, and another at the variable level to account for the influence of each variable. In order to retain interpretability, reproduce clinical activities, and integrate sequential information, RETAIN was developed as a reverse attention mechanism. Predictive models for heart failure and cataract risk have been suggested by Kwon and colleagues [145] based on RETAIN (RetainVis). DL models' interpretability may be improved by identifying certain positions (e.g., DNA, visits, time) in a sequence (e.g., DNA, visits, time) where those input attributes might impact the output of prediction. Since 2016, XAI researchers have been studying the complementary and alternative medicine (CAM) [146] approach and its modifications [147], which have since been put to use in digital healthcare, particularly in medical picture analysis. According to the authors of [148], a limited dataset of CT scans was utilized for training Grad-CAM to detect appendicitis. To diagnose hypoglycemia, Porumb et al. [149] used a combination of RNN and CNN for an electrocardiogram (ECG) experiment with Grad-CAM to identify the heartbeat segments that were most relevant. The coronavirus classification method was applied with a multiscale camera to identify the diseased regions in a work [150]. These saliency maps are suggested because of their ease of visual interpretation. Clinical end users are not given precise responses by attention-based XAI approaches, however, these methods do indicate regions of increased concern, making it simpler for them to make decisions. However, the primary drawbacks, such as information overload and warning fatigue. It may not be desirable to highly convey this information to a clinical end user.

### E. Surrogate representation using XAI

Individual health-related factors that lead to disease prediction are effectively recognized by using XAI in the medical field by applying the local interpretable model-agnostic explanation (LIME) method [66]. This method provides explanations for any classifier by approximating it with a surrogate interpretable and locally faithful representation. Linear models in the neighborhood are learned by LIME in order to explain the disruption of an instance [151]. LIME was utilized by Pan et al. [152] to examine the impact of additional cases on the prediction of central precocious puberty in children. When it comes to AI survival models, Kovalev et al. [153] came up with an approach called SurvLIME to describe it. According to the authors of [154], local post-hoc explanation model was applied in their investigation of the segmented lung suspicious items. An AI system built by Panigutti and his colleagues [155] can forecast the patient's return to the hospital as well as their diagnosis and prescription order. LIME's implemented system uses a rule-based explanation to train a local surrogate model, which may be mined employing a decision tree with many labels. This system is similar to LIME's implementation. For the purpose of predicting acute critical illness via the use of EHRs, Lauritsen et al. [156] investigated an XAI technique employing Layer-wise Relevance Propagation [157]. The white-box estimation needs to precisely reflect the black-box algorithm in order to provide a trustworthy description, which is why XAI uses surrogate representation so often. Comprehension by the physician may be compromised if the surrogate algorithms are very complex. Table II depicts the overview of the above methods. In addition, Table III presents different XAI algorithms and Table IV describes particularly SHAP XAI algorithm reported in the literature.

TABLE II

XAI APPROACHES USED IN MEDICINE AND HEALTHCARE AS WELL AS THEIR TYPES

| Type | Method | Post-hoc | Intrinsic | Local | Global | Application | Ref. |
|---|---|---|---|---|---|---|---|
| Dimension reduction using XAI | Optimum feature selection | ✗ | ✓ | ✗ | ✓ | Prediction of enhancers that are cell type specific | [109] |
| | Laplacian Eigenmaps | ✗ | ✓ | ✗ | ✓ | Classification of Brain tumor from MRI | [107] |
| | Sparse-balanced SVM | ✗ | ✓ | ✗ | ✓ | Diagnosis of type 2 diabetes at an early stage | [158] |
| | Cluster analysis and LASSO | ✗ | ✓ | ✗ | ✓ | Classification of lung cancer patients | [108] |
| | Sparse DL | ✗ | ✓ | ✗ | ✓ | Glioblastoma multiforme long-term survival prediction | [110] |
| | Optimal feature selection | ✗ | ✓ | ✗ | ✓ | Estimation of side effect for drug | [106] |
| Rule extraction and knowledge | Decision rules | ✗ | ✓ | ✗ | ✓ | Stroke Prediction | [126] |
| | Rule-based system | ✗ | ✓ | ✗ | ✓ | Forecasts for 30-day readmissions and pneumonia risk. | [14] |
| | Mimic learning | ✓ | ✗ | ✓ | ✓ | Predicting outcomes in the ICU for severe lung injury | [120] |

| Type | Method | Post-hoc | Intrinsic | Local | Global | Application | Ref. |
|---|---|---|---|---|---|---|---|
| distillation using XAI | Textual or visual justification | ✓ | ✗ | ✓ | ✓ | Classification of breast mass | [125] |
| | Visualization rules | ✓ | ✗ | ✗ | ✓ | Clinical diagnosis of diabetes and breast cancer | [121] |
| | Lists of Bayesian rule | ✗ | ✓ | ✗ | ✓ | Stroke prediction | [120] |
| | Fuzzy rules | ✗ | ✓ | ✗ | ✓ | Prediction of in-hospital mortality for all cases | [123] |
| Feature importance or selection using XAI | DeepLIFT | ✓ | ✗ | ✗ | ✓ | Detection of splice site | [133] |
| | Feature weighting | ✓ | ✗ | ✗ | ✓ | Prediction of ICU mortality for all-causes | [128] |
| | DeepLIFT | ✓ | ✗ | ✓ | ✓ | Ophthalmic diagnosis | [137] |
| | Feature marginalization | ✓ | ✗ | ✓ | ✓ | Diagnosis of inflammatory or microbiota bowel diseases from skin and gut | [131] |
| Attention mechanism in XAI | Attention | ✗ | ✓ | ✓ | ✗ | Prediction of future hospitalization from EHR | [143] |
| | | ✗ | ✓ | ✓ | ✗ | ICU clinical events predictions | [140] |
| | | ✗ | ✓ | ✓ | ✗ | HIV genome integration site prediction | [142] |
| | | ✗ | ✓ | ✓ | ✓ | Prediction of clinical risk for cardiac cataract or failure | [145] |
| | | ✗ | ✓ | ✓ | ✗ | Prediction of heart failure | [144] |
| | MLCAM | ✗ | ✓ | ✓ | ✗ | Localization of brain tumor | [145] |
| | Grad-CAM | ✗ | ✓ | ✓ | ✗ | Appendicitis diagnosis | [147] |
| | Grad-CAM | ✗ | ✓ | ✓ | ✗ | Hypoglycaemia detection using ECG | [149] |
| Surrogate representation using XAI | LIME | ✓ | ✗ | ✓ | ✗ | Prediction of early puberty in the central region | [152] |
| | | ✓ | ✗ | ✓ | ✗ | Construction of survival models | [153] |
| | | ✓ | ✗ | ✓ | ✗ | Diagnosis of autism spectrum disorder | [159] |
| | Rule-based XAI | ✓ | ✗ | ✓ | ✗ | Patient medications, diagnosis and readmission prediction | [155] |

TABLE III

APPLICATIONS OF XAI ALGORITHMS IN DIFFERENT SCENARIOS

| Algorithm | Ref. | Location | Mode | Remarks |
|---|---|---|---|---|
| LRP | [175] | Brain | MRI | employs LRP to locate Alzheimer's disease-causing brain areas, where LRP is proved to be more specific than guided backpropagation in detecting the disease areas. |
| Multiple instance learning | [176] | Dental Eye | Fundus Image | Presents a cutting-edge DL-based diabetic retinopathy grading system based on a medically comprehensible explanation; and infers an image grade. |
| Trainable attention | [177] | Dental Eye | Fundus Image | proposes an attention-based CNN for removing high redundancy in fundus images for glaucoma detection. |
| Trainable attention | [178] | Chest | X-ray | Proposes two novel neural networks for the detection of chest radiographs containing pulmonary lesions; both the architectures make use of a large number of weakly-labelled images combined with a smaller number of manually annotated X-rays. |
| Grad-CAM , LIME | [179] | Chest | X-ray | Presents several examples of CXRs with pneumonia infection, together with the results from the grad-CAM and LIME; LIME's explanatory output is presented as the superpixels with the highest positive weights superimposed on the original input. |
| CAM | [180] | Prostate | Histology | constructs a patch-wise predictive model using convolutional neural networks to detect cancerous Gleason patterns, and examines the interpretability of trained models using Class Activation Maps (CAMs). |
| Saliency propagation | [181] | Skin | Vitiligo image | Proposes a unique weakly supervised approach to segmenting vitiligo lesions with preserved fine borders; the activation map can be paired with superpixel-based saliency propagation to generate segmentation with well-preserved edges. |
| Guided backpropagation | [182] | Gastrointestinal | Endoscopy | Uses Guided Backpropagation based Fully Convolutional Networks to investigate the pixels necessary for identifying polyps (FCNs). |
| TCAV | [183] | Cardiovascular | MRI | using TCAV (Testing with Concept Activation Vectors) to show which biomarkers are already known to be linked to cardiac disease. |
| TCAV | [184] | Breast and Eye | histopathology and retinopathy | proposing the addition of regression concept vector to TCAV in identifying tumor size, while TCAV represented binary ideas, regression concept vectors indicated continuous-valued measurements. |

TABLE IV

DISCUSSION OF SHAP XAI ALGORITHMS IN THE LITERATURE

| Ref | Disease Diagnosis | ML Models Used | Work Summary |
|---|---|---|---|
| [185] | Chronic Obstructive Pulmonary Disease | Gradient boosting machine (GBM) | The work ranks the attributes crucial to the GBM model according to their average absolute SHAP values that reflect the impact of each feature on a prediction. |
| [186] | Parkinson's disease diagnosis | deep forest (gcForest), extreme gradient boosting (XGBoost), light gradient boosting machine (LightGBM) and random forest (RF) | The classifiers determine how much the SHAP value contributes to the features. When 150 features are considered, SHAP-gcForest achieves 91.78 percent classification accuracy. With 50 features, SHAP-LightGBM combined with LightGBM yields 91.62% accuracy. |
| [187] | predicting acute kidney injury progression | XGboost , Logistic Regression (LR) | The SHAP value increases as creatinine levels climb until it reaches about 5 mg/dL. As the Furosemide Stress Test (FST) is increased until it reaches 100 ml/min, the SHAP result decreases. |

| [188] | COVID-19 Prediction | XGboost model, LGBM, gradient boosting classifier (GBC), categorical boosting (CatBoost), RF | Harris hawks optimization (HHO) approach is employed to fine-tune the hyperparameters of a few cutting-edge classifiers, including the HHO-RF, HHO-XGB, HHO-LGB, and ensemble methods, in order to increase classification accuracy. |
|---|---|---|---|
| [189] | Prediction of COVID-19 diagnosis based on symptoms | GBM | A gradient-boosting machine model created with decision-tree base-learners was used to produce predictions. SHAP values calculate the contribution of each feature to overall model predictions by average across samples. |
| [190] | Efficient analysis of COVID-19 | Support Vector Machine (SVM), Naive Bayes (NB), Multiple Linear Regression (MLP), K-Nearest Neighbors (KNN), RF, LR, and Decision Tree (DT), Keras Classifier | Simple classification techniques like random forest can outperform keras with Boruta for feature selection when the dataset size is not too large. Computing the information gain values for each attribute in Clinical Data1 shows the importance of each attribute. |
| [191] | COVID-19 Diagnosis | SqueezeNet, LIME | LIME and SHAP are compared for COVID diagnosis using SqueezeNet to recognize pneumonia, COVID-19, and normal lung imaging. Results show LIME and SHAP can boost the model's transparency and interpretability. |
| [192] | COVID-19 vaccine prioritization | Random Forest, XGBoost classifiers, XAI | CovidXAI predicts a user's risk group using Random Forest and XGBoost classifiers. CovidXAI uses 24 criteria to define an individual's risk category and vaccine urgency. |
| [193] | COVID-19 Pneumonia Classification | XGBoost, Random Forest | The goal is to provide grounds for understanding the distinctive COVID-19 radiographic texture features using supervised ensemble ML methods based on trees through the interpretable SHAP approach. SHAP recursive feature elimination with cross-validation is used to select features. The best classification model was XGBoost, with an accuracy of 0.82 and a sensitivity of 0.82. |
| [194] | Lung cancer hospital length of stay prediction | Random Forest, logistic regression | This paper introduces a predictive LOS framework for lung cancer patients utilizing ML models. Using SHAP, the output of the predictive model (RF) with SMOTE class balancing approach is understood demonstrating the most relevant clinical factors that contributed to predicting lung cancer LOS using the RF model. |

## VI. DISCUSSION ON DIFFERENT ASPECTS OF XAI

It is still a work in progress to design and govern AI systems that can be trusted. We looked at the function that explainability plays in constructing trustworthy AI since the two concepts are intertwined. A framework with actual suggestions for choosing between classes of XAI approaches is provided by us as an extension to previous recent surveys [12, 19, 85, 92, 113]. As a result, we presented useful definitions and made contributions to the literature on quantitative assessment measures. XAI is being discussed as a part of a broader effort to build confidence in AI.

Much work on automated diagnosis and prognosis has been done in the medical area using ML techniques [29]. We can see from grand-challenge.org that a number of medical issues have surfaced and energized ML and AI researchers. DL models [160], [161] that use U-Net for medical segmentation are effective. U-Net, on the other hand, is still a mystery since it is a DL neural network. There are several more approaches and special transformations (denoising, for example) that have been described in MICCAI papers, such as [162]. The issue of interpretability in medicine goes much beyond idle speculation. Medical interpretability contains risks and duties that other areas do not take into consideration [67]. There is a risk of death associated with medical decisions that are made. The pursuit of medical explainability has therefore resulted in several further publications [71]. To raise awareness about the significance of interpretability in the medical sector, they summarize prior studies [163], consisting of subfield-specific reviews, i.e., [74] for sentiment analysis and chest radiography in medicine [75]. As stated explicitly in [60], AI in dermatology is severely limited because it cannot deliver individualized assessments [60].

### A. Developing XAI to create trustworthy AI

The reason for requiring explainability is that it dictates what ought to be communicated as well as a step-by-step guide for developers with real diagram suggestions was offered. The development of post-hoc explanation techniques that contain argumentation evidence for their assertions [164] might be a solution to the problem of misleading post-hoc explanations. Model-based explanations can be suitable when looking for a post-hoc explanation since they are full and contain quantitative proxy criteria to assess their soundness. Many more studies are required to better understand the performance of explainable models in the healthcare industry [165] and to develop explainable modeling approaches such as rule-based systems or GAMs with or without interaction terms. Thus, the importance of feature engineering in the context of feature interpretation is further highlighted. Another potential research approach is to create hybrid techniques that combine knowledge-driven and data-driven aspects for feature engineering or feature selection. The quality of explanations is critical when utilizing XAI to build trustworthy AI. According to Poursabzi-Sangdeh et al. [84], improved results from human-machine tasks are not always accompanied by greater performance. We discovered that evaluating the clarity of local explanations is challenging and that there are currently no quantitative assessment measures for example-based methodologies. Even while interpretability is commonly recognized to be user-dependent, it has not been measured as such when assessing the quality of explanations [85]. Future research should focus on establishing a criterion for identifying when AI systems can be explained to a wide range of people.

### B. Complementary approaches to the development of trustworthy healthcare AI

In some cases (e.g. [165]), explanations are not essential nor adequate to build faith in AI. Perceived system capabilities, control, and predictability are other crucial factors [166]. Because of this, even while XAI has the potential to contribute to reliable AI, it is not without its limitations [167]. The following are some additional approaches that may be implemented in conjunction with AI in healthcare to ensure its trustworthiness:

The first one is regarding the quality of data. It is important for providing information on the quality of the data. Real-world data may include biases, inaccuracies, or be incomplete since it was not gathered for study reasons. This means that knowing the quality of the data obtained and how it was acquired is at least as crucial to the final model's constraints as its explainability [168]. Kahn et al. [169] provide a generally recognized approach for determining whether EHR data is appropriate for a given use case. The second aspect is about conducting comprehensive testing. The use of standard data formats in health care has made external validation more viable, even if it is still considered an area for the development of models of clinical risk prediction [87, 170]. Stability, fairness, and privacy are only a few of the topics being studied in this field. Another aspect is the governing law.

*C. Trends in Existing Research*

Some interesting trends are visible in the works reported in the literature. These studies have been described in detail in the previous section, and now some of the insights are described in the following. Literature indicates post-hoc and model-specific explanations can be used to categorize XAI model explanations. Gradient-based, perturbation-based, and rule-based attribution are three different methods. XAI can assist in explaining how the reduced features are connected to the original variables and how they impact the model's performance in the context of dimension reduction. Laplacian Eigenmaps, ICA, and PCA techniques for dimension reduction utilizing XAI are available. Other similar techniques are feature importance, and partial dependence plots. Studies so far indicate that in the field of XAI, rule extraction and knowledge distillation are two crucial concepts. The permutation importance method is one of the most widely used XAI strategies for feature importance. Another XAI technique for determining feature importance is the SHAP, partial dependence plots, and methods based on decision trees, including the Gini index and information gain. An electrocardiogram experiment with Grad-CAM can be used to diagnose hypoglycemia by combining RNN and CNN. In XAI, examples of surrogate models found in the literature are decision trees and linear regression models.

Several XAI techniques are difficult to compute with and may be challenging to scale to large datasets or complex models. There is still a shortage of XAI methods interaction with useful systems. There is not enough information on how humans interpret and apply XAI explanations. However, some recent studies discuss the idea of federated learning of XAI models, which was created specifically to fulfill these two criteria simultaneously. Recent studies also consider the application of XAI in the field of medical data standardization which is a still a challenging issue in many underdeveloped or developing countries. Moreover, some studies are focused to developing hybrid approaches of XAI where individual XAI techniques are combined for better performance. Furthermore, some recent works are concentrating on research intersecting quantum computing and XAI as quantum computing has the potential to improve the performance of XAI for healthcare.

*D. Challenges and Future Prospects*

Intractable systems may be simplified using models. A lack of utilization of DNNs with a large number of parameters may result in under-utilized ML capabilities. It is feasible to take advantage of the excitement around predictive science by examining if it is viable to add new, more complex components to an existing model, such that sections of these components may be recognized as fresh insights. The enhanced model must, of course, be similar to earlier demonstrations and models, with a clear interpretation of why the additional components correlate to formally ignored insights.

Some approaches that are deeply embedded in particular industries are still widely used, while powerful ML algorithms are finding new uses. Fragmented and experimental implementations of existing or custom-made interpretable techniques are common in medical ML, which is still in its infancy. We may still have large undiscovered prospects in interpretability research despite the current emphasis on increasing accuracy and performance in feature selection and extraction.

## VII. CONCLUSIONS

This review paper summarizes the state-of-the-art, evaluates the level of quality of recent research, identifies the research gap, and makes recommendations for future studies that can significantly develop XAI. In particular, this review focuses on the application of XAI techniques to the healthcare industry. Discussions are provided on the importance of using XAI, the scenarios where XAI is suitable, and the methods of implementing XAI. In this regard, this paper provides an overview of the literature, connecting various perspectives and advocating specific design solutions. It is discussed here that AI has the potential to alter medical practice and the role of physicians. Moreover, explainable models may be favored over posthoc explanations when XAI is used to develop trustworthy AI for healthcare. Rigorous regulation and validation may be required in addition to reporting data quality and extensive validation, as limited evidence supports the practical utility of explainability. In the future, efforts will be necessary to address the problems and concerns that have been raised with XAI.

APPENDIX

((''Explainable AI'') OR (''Interpretable AI'') OR (''Explainable Artificial Intelligence'') OR (''Interpretable Artificial Intelligence'') OR (''XAI'')) AND ((''medical'') OR (''healthcare''))

ACKNOWLEDGMENT

This work has been carried out at the Institute of Information

and Communication Technology (IICT) of Bangladesh University of Engineering and Technology (BUET), Bangladesh. Hence, the authors would like to thank BUET for providing the support.

REFERENCES

[1] A. Gupta, A. Anpalagan, L. Guan, and A. S. Khwaja, "Deep learning for object detection and scene perception in self-driving cars: Survey, challenges, and open issues," *Array,* vol. 10, p. 100057, 2021.

[2] S. Bharati, P. Podder, and M. R. H. Mondal, "Artificial Neural Network Based Breast Cancer Screening: A Comprehensive Review," *International Journal of Computer Information Systems and Industrial Management Applications,* vol. 12, pp. 125-137, 2020.

[3] M. R. H. Mondal, S. Bharati, and P. Podder, "CO-IRv2: Optimized InceptionResNetV2 for COVID-19 detection from chest CT images," *PloS one,* vol. 16, no. 10, p. e0259179, 2021.

[4] S. Bharati, P. Podder, and M. R. H. Mondal, "Hybrid deep learning for detecting lung diseases from X-ray images," *Informatics in Medicine Unlocked,* p. 100391, 2020.

[5] A. Rajkomar *et al.*, "Scalable and accurate deep learning with electronic health records," *NPJ Digital Medicine,* vol. 1, no. 1, pp. 1-10, 2018.

[6] S. Tonekaboni, S. Joshi, M. D. McCradden, and A. Goldenberg, "What clinicians want: contextualizing explainable machine learning for clinical end use," presented at the Machine learning for healthcare conference, 2019, 2019.

[7] E. D. Peterson, "Machine learning, predictive analytics, and clinical practice: can the past inform the present?," *Jama,* vol. 322, no. 23, pp. 2283-2284, 2019.

[8] J. He, S. L. Baxter, J. Xu, J. Xu, X. Zhou, and K. Zhang, "The practical implementation of artificial intelligence technologies in medicine," *Nature medicine,* vol. 25, no. 1, pp. 30-36, 2019.

[9] Z. C. Lipton, "The Mythos of Model Interpretability: In machine learning, the concept of interpretability is both important and slippery," *Queue,* vol. 16, no. 3, pp. 31-57, 2018.

[10] J. Burrell, "How the machine 'thinks': Understanding opacity in machine learning algorithms," *Big Data & Society,* vol. 3, no. 1, p. 2053951715622512, 2016.

[11] F. Doshi-Velez and B. Kim, "Considerations for evaluation and generalization in interpretable machine learning," in *Explainable and interpretable models in computer vision and machine learning*: Springer, 2018, pp. 3-17.

[12] E. J. Hwang *et al.*, "Development and validation of a deep learning–based automated detection algorithm for major thoracic diseases on chest radiographs," *JAMA network open,* vol. 2, no. 3, pp. e191095-e191095, 2019.

[13] J. G. Nam *et al.*, "Development and validation of deep learning–based automatic detection algorithm for malignant pulmonary nodules on chest radiographs," *Radiology,* vol. 290, no. 1, pp. 218-228, 2019.

[14] K. J. Geras *et al.*, "High-resolution breast cancer screening with multi-view deep convolutional neural networks," *arXiv preprint arXiv:1703.07047,* 2017.

[15] A. Khamparia *et al.*, "Diagnosis of Breast Cancer Based on Modern Mammography using Hybrid Transfer Learning," *Multidimensional Systems and Signal Processing,* 2020.

[16] M. R. H. Mondal, S. Bharati, and P. Podder, "Diagnosis of COVID-19 Using Machine Learning and Deep Learning: A Review," *Current Medical Imaging,* 2021.

[17] S. Bharati, P. Podder, M. Mondal, and V. B. Prasath, "Medical Imaging with Deep Learning for COVID-19 Diagnosis: A Comprehensive Review," *International Journal of Computer Information Systems and Industrial Management Applications,* vol. 13, pp. 91 - 112, 2021.

[18] E. J. Topol, "High-performance medicine: the convergence of human and artificial intelligence," *Nature medicine,* vol. 25, no. 1, pp. 44-56, 2019.

[19] A. B. Arrieta *et al.*, "Explainable Artificial Intelligence (XAI): Concepts, taxonomies, opportunities and challenges toward responsible AI," *Information fusion,* vol. 58, pp. 82-115, 2020.

[20] R. Basnet, M. O. Ahmad, and M. N. S. Swamy, "A deep dense residual network with reduced parameters for volumetric brain tissue segmentation from MR images," *Biomedical Signal Processing and Control,* vol.70, p. 103063, 2021.

[21] S. Bharati and M. R. Hossain Mondal, "12 Applications and challenges of AI-driven IoHT for combating pandemics: a reviewComputational Intelligence for Managing Pandemics," A. Khamparia, R. Hossain Mondal, P. Podder, B. Bhushan, V. H. C. d. Albuquerque, and S. Kumar, Eds.: De Gruyter, 2021, pp. 213-230.

[22] P. Podder, S. Bharati, M. R. H. Mondal, and U. Kose, "Application of Machine Learning for the Diagnosis of COVID-19," in *Data Science for COVID-19*: Elsevier, 2021, pp. 175-194.

[23] E. Jabason, M. O. Ahmad, and M. N. S. Swamy, "Classification of Alzheimer's Disease from MRI Data Using a Lightweight Deep Convolutional Model," In *2022 IEEE International Symposium on Circuits and Systems (ISCAS)*, pp. 1279-1283. IEEE, 2022.

[24] S. Bharati, P. Podder, M. Mondal, and V. B. Prasath, "CO-ResNet: Optimized ResNet model for COVID-19 diagnosis from X-ray images," *International Journal of Hybrid Intelligent Systems,* vol. 17, pp. 71-85, 2021.

[25] M. R. H. Mondal, S. Bharati, P. Podder, and P. Podder, "Data analytics for novel coronavirus disease," *Informatics in Medicine Unlocked,* vol. 20, p. 100374, 2020.

[26] G. Yang, Q. Ye, and J. Xia, "Unbox the black-box for the medical explainable AI via multi-modal and multi-centre data fusion: A mini-review, two showcases and beyond," *Information Fusion,* vol. 77, pp. 29-52, 2022/01/01/ 2022.

[27] B. Mahbooba, M. Timilsina, R. Sahal, and M. Serrano, "Explainable artificial intelligence (xai) to enhance trust management in intrusion detection systems using decision tree model," *Complexity,* vol. 2021, 2021.

[28] S. R. Islam, W. Eberle, and S. K. Ghafoor, "Towards quantification of explainability in explainable artificial intelligence methods," In *The thirty-third international flairs conference* 2020, 2020.

[29] F. Jiang *et al.*, "Artificial intelligence in healthcare: past, present and future," *Stroke and vascular neurology,* vol. 2, no. 4, pp. 230-243, 2017.

[30] T. Panch, H. Mattie, and L. A. Celi, "The "inconvenient truth" about AI in healthcare," *NPJ digital medicine,* vol. 2, no. 1, pp. 1-3, 2019.

[31] K.-H. Yu, A. L. Beam, and I. S. Kohane, "Artificial intelligence in healthcare," *Nature biomedical engineering,* vol. 2, no. 10, pp. 719-731, 2018.

[32] D. Shen, G. Wu, and H.-I. Suk, "Deep learning in medical image analysis," *Annual review of biomedical engineering,* vol. 19, pp. 221-248, 2017.

[33] G. Litjens *et al.*, "A survey on deep learning in medical image analysis," *Medical image analysis,* vol. 42, pp. 60-88, 2017.

[34] J. Ker, L. Wang, J. Rao, and T. Lim, "Deep learning applications in medical image analysis," *Ieee Access,* vol. 6, pp. 9375-9389, 2017.

[35] S. E. Dilsizian and E. L. Siegel, "Artificial intelligence in medicine and cardiac imaging: harnessing big data and advanced computing to provide personalized medical diagnosis and treatment," *Current cardiology reports,* vol. 16, no. 1, p. 441, 2014.

[36] V. L. Patel *et al.*, "The coming of age of artificial intelligence in medicine," *Artificial intelligence in medicine,* vol. 46, no. 1, pp. 5-17, 2009.

[37] S. Jha and E. J. Topol, "Adapting to artificial intelligence: radiologists and pathologists as information specialists," *Jama,* vol. 316, no. 22, pp. 2353-2354, 2016.

[38] E. Strickland, "IBM Watson, heal thyself: How IBM overpromised and underdelivered on AI health care," *IEEE Spectrum,* vol. 56, no. 4, pp. 24-31, 2019.

[39] N. S. Weingart, R. M. Wilson, R. W. Gibberd, and B. Harrison, "Epidemiology of medical error," *Bmj,* vol. 320, no. 7237, pp. 774-777, 2000.

[40] M. L. Graber, N. Franklin, and R. Gordon, "Diagnostic error in internal medicine," *Archives of internal medicine,* vol. 165, no. 13, pp. 1493-1499, 2005.

[41] I. Doroniewicz *et al.*, "Computer-based analysis of spontaneous infant activity: A pilot study," in *Information Technology in Biomedicine*: Springer, 2021, pp. 147-159.

A Review on Explainable Artificial Intelligence for Healthcare 15[169] M. G. Kahn *et al.*, "A harmonized data quality assessment terminology and framework for the secondary use of electronic health record data," *Egems,* vol. 4, no. 1, 2016.

[170] B. A. Goldstein, A. M. Navar, M. J. Pencina, and J. Ioannidis, "Opportunities and challenges in developing risk prediction models with electronic health records data: a systematic review," *Journal of the American Medical Informatics Association,* vol. 24, no. 1, pp. 198-208, 2017.

[171] V. Vishwarupe, P. M. Joshi, N. Mathias, S. Maheshwari, S. Mhaisalkar, V. Pawar, "Explainable AI and Interpretable Machine Learning: A Case Study in Perspective," Procedia Computer Science, Volume 204, 2022, Pages 869-876.

[172] T. Miller, Explanation in Artificial Intelligence: Insights from the Social Sciences. Artif. Intell. 2019, 267, 1–38.

[173] A. Adadi, M. Berrada, "Peeking Inside the Black-Box: A Survey on Explainable Artificial Intelligence (XAI)," IEEE Access 2018, 6, 52138–52160.

[174] L.H. Gilpin, D. Bau, B. Z. Yuan, A. Bajwa, M. Specter, L. Kagal, "Explaining Explanations: An Overview of Interpretability of Machine Learning," In *Proceedings of the 2018 IEEE 5th International Conference on Data Science and Advanced Analytics (DSAA),* Turin, Italy, 1–4 October 2018; pp. 80–89.

[175] M. Böhle, et al., "Layer-wise relevance propagation for explaining deep neural network decisions in MRI-based Alzheimer's disease classification," *Frontiers in aging neuroscience* 11 (2019): 194.

[176] T. Araújo et al., "DR|GRADUATE: Uncertainty-aware deep learning-based diabetic retinopathy grading in eye fundus images," *Medical Image Analysis* 63 (2020): 101715.

[177] L. Li et al., "A large-scale database and a CNN model for attention-based glaucoma detection," *IEEE transactions on medical imaging* 39.2 (2019): 413-424.

[178] E. Pesce et al., "Learning to detect chest radiographs containing pulmonary lesions using visual attention networks," *Medical image analysis* 53 (2019): 26-38.

[179] S. Rajaraman et al., "Visualizing and explaining deep learning predictions for pneumonia detection in pediatric chest radiographs," Medical Imaging 2019: Computer-Aided Diagnosis. Vol. 10950. SPIE, 2019.

[180] J. Silva-Rodríguez et al., "Going deeper through the Gleason scoring scale: An automatic end-to-end system for histology prostate grading and cribriform pattern detection," *Computer Methods and Programs in Biomedicine* 195 (2020): 105637.

[181] Z. Bian et al., "Weakly supervised Vitiligo segmentation in skin image through saliency propagation," 2019 *IEEE International Conference on Bioinformatics and Biomedicine (BIBM)*. IEEE, 2019.

[182] K. Wickstrøm, M. Kampffmeyer, and R. Jenssen, "Uncertainty and interpretability in convolutional neural networks for semantic segmentation of colorectal polyps," *Medical image analysis* 60 (2020): 101619.

[183] J. R. Clough et al., "Global and local interpretability for cardiac MRI classification," *International Conference on Medical Image Computing and Computer-Assisted Intervention*. Springer, Cham, 2019.

[184] M. Graziani et al., "Concept attribution: Explaining CNN decisions to physicians," *Computers in biology and medicine* 123 (2020): 103865.

[185] C.-T. Kor, Y.-R. Li, P.-R. Lin, S.-H. Lin, B.-Y. Wang, and C.-H. Lin, "Explainable Machine Learning Model for Predicting First-Time Acute Exacerbation in Patients with Chronic Obstructive Pulmonary Disease," Journal of Personalized Medicine, vol. 12, no. 2, p. 228, Feb. 2022, doi: 10.3390/jpm12020228.

[186] Y. Liu, Z. Liu, X. Luo, H. Zhao, "Diagnosis of Parkinson's disease based on SHAP value feature selection," Biocybernetics and Biomedical Engineering, Volume 42, Issue 3, 2022, Pages 856-869.

[187] C. Wei, L. Zhang, Y. Feng et al., "Machine learning model for predicting acute kidney injury progression in critically ill patients," *BMC Med Inform Decis Mak* 22, 17 (2022).

[188] K. Debjit et al., "An Improved Machine-Learning Approach for COVID-19 Prediction Using Harris Hawks Optimization and Feature Analysis Using SHAP," *Diagnostics*, vol. 12, no. 5, p. 1023, Apr. 2022.

[189] Y. Zoabi, , S. Deri-Rozov, N. Shomron, "Machine learning-based prediction of COVID-19 diagnosis based on symptoms," *npj Digit. Med.* 4, 3 (2021).

[190] S. Ali, Y. Zhou, M. Patterson, "Efficient analysis of COVID-19 clinical data using machine learning models," *Med Biol Eng Comput* 60, 1881–1896 (2022).

[191] J. H. Ong, K. M. Goh and L. L. Lim, "Comparative Analysis of Explainable Artificial Intelligence for COVID-19 Diagnosis on CXR Image," 2021 *IEEE International Conference on Signal and Image Processing Applications*, 2021, pp. 185-190.

[192] D. Chowdhury, S. Poddar, S. Banarjee et al., "CovidXAI: explainable AI assisted web application for COVID-19 vaccine prioritization," *Internet Technology Letters*, 2022; 5(4):e381.

[193] L. V. De Moura et al., "Explainable Machine Learning for COVID-19 Pneumonia Classification With Texture-Based Features Extraction in Chest Radiography," *Frontiers in digital health* 3 (2021).

[194] B. Alsinglawi, O .Alshari, M. Alorjani et al., "An explainable machine learning framework for lung cancer hospital length of stay prediction," *Sci Rep* 12, 607 (2022).

[195] A. F. Markus, J. A. Kors, and P. R. Rijnbeek, "The role of explainability in creating trustworthy artificial intelligence for health care: a comprehensive survey of the terminology, design choices, and evaluation strategies," *Journal of Biomedical Informatics*, vol. 113, 2021.
## AUTHORS' BIOGRAPHIES

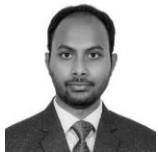

**Subrato Bharati** was with the Institute of Information and Communication Technology, BUET, Bangladesh, and now with the Department of Electrical and Computer Engineering, Concordia University, Montreal, QC, Canada. He authored/co-authored over 50 Journal articles, Conference Proceedings, and Book Chapters published by IEEE, Elsevier, Springer, and others. He is an Associate Editor of Journal of the International Academy for Case Studies and a Guest Editor of a special issue in Journal of Internet Technology. His research interests include health informatics, medical imaging, digital signal and image processing, eXplainable AI, machine and deep learning.

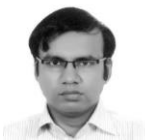

**M. Rubaiyat Hossain Mondal** obtained the PhD degree in 2014 from the Department of Electrical and Computer Systems Engineering, Monash University, Melbourne, Australia. From 2005 to 2010, and from 2014 to date, he has been working as a Faculty Member at the Institute of Information and Communication Technology (IICT) in BUET, Bangladesh. He has published a number of papers in journals, conferences and book chapters, and edited several books published by reputed publishers. His research interests include artificial intelligence, image processing, bioinformatics, wireless communications, and cryptography.

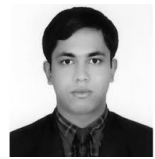

**Prajoy Podder** is currently a Researcher at the Institute of Information and Communication Technology, Bangladesh University of Engineering and Technology. He authored/co-authored over 45 Journal articles, Conference Proceedings, and Book Chapters published by IEEE, Elsevier, Springer, and others. His research interests include Digital Image Processing, Data Mining, IoT, Machine Learning, and Health Informatics.